\documentclass{article}
\usepackage{ijcai21}

\usepackage{times}

\usepackage{soul}
\usepackage{url}
\usepackage[hidelinks]{hyperref}
\usepackage[utf8]{inputenc}
\usepackage[small]{caption}
\usepackage{graphicx}
\usepackage{amsmath}
\usepackage{booktabs}
\usepackage[para,online,flushleft]{threeparttable}
\usepackage{makecell}
\usepackage{multirow}
\usepackage{bm}
\usepackage{algorithm} 
\usepackage{algorithmic}
\usepackage[algo2e]{algorithm2e} 

\usepackage{adjustbox}
\usepackage{color}

\title{Towards Fast and Accurate Multi-Person Pose Estimation on Mobile Devices}

\author{
Xuan Shen\thanks{These Authors contributed equally.}$^1$\and
Geng Yuan\text{$^*$}$^1$\and
Wei Niu$^2$\and
Xiaolong Ma$^1$\and\\
Jiexiong Guan$^2$\and
Zhengang Li$^1$\and
Bin Ren$^2$\And
Yanzhi Wang$^1$
\affiliations
$^1$Northeastern University, \\
$^2$College of William \& Mary \\
\emails
\{shen.xu, yuan.geng, ma.xiaol, li.zhen, yanz.wang\}@northeastern.edu,\\
\{wniu, jguan\}@email.wm.edu, 
bren@cs.wm.edu,
}
\begin{document}

\maketitle

\begin{abstract}
The rapid development of autonomous driving, abnormal behavior detection, and behavior recognition makes an increasing demand for multi-person pose estimation-based applications, especially on mobile platforms. However, to achieve high accuracy, state-of-the-art methods tend to have a large model size and complex post-processing algorithm, which costs intense computation and long end-to-end latency. To solve this problem, we propose an architecture optimization and weight pruning framework to accelerate inference of multi-person pose estimation on mobile devices. With our optimization framework, we achieve up to 2.51$\times$ faster model inference speed with higher accuracy compared to representative lightweight multi-person pose estimator.

\end{abstract}

\newcommand{\GY}[1]{\textcolor{blue}{Geng: #1}}

\newcommand{\todo}[1]{\textcolor{red}{\sf\bfseries Todo: #1}}

\newcommand{\red}[1]{\textcolor{red}{#1}}

\newcommand{\blue}[1]{\textcolor{blue}{#1}}

\section{Introduction}
In recent years, with the increasing popularity of mobile computing and edge AI, mobile platforms have become an important carrier of DNN applications~\cite{li2021real,zhao2020achieving,niu2020achieving}.
multi-person pose estimation (PE) is one of the popular DNN applications that play an important role in many fields, including autonomous driving, abnormal behavior detection, and behavior recognition, and has made impressive progress with the joint efforts. 

The multi-person PE can be achieved mainly in two design methodologies, which are the top-down method and the bottom-up method.
The top-down method~\cite{fang2017rmpe} performs PE on a single person who is captured by a detector in advance.
The accuracy will be dependent on the detector which might fail to find all the persons when they are close together.
And the overall inference time will increase with the increasing number of people in an image frame, which is not efficient for multi-person PE.
On the other hand, the bottom-up method~\cite{openpose2019} predicts the human keypoints for all people in an image frame at the same time.
The bottom-up methods usually have smaller model sizes and fewer computation counts (i.e., GFLOPs), making the bottom-up methods more desirable for fast multi-person PE.
However, the state-of-the-art bottom-up methods are still computational intensive, usually requiring tens to hundreds of GFLOPs. This is challenging for fast inference, especially on resource-limited mobile devices.

DNN weight pruning~\cite{he2019filter,li2021npas,zhang2021clicktrain}, as an effective model compression technique, has been widely adopted to remove redundant weights in DNN models, reducing the required resources for both model storage and computation, accelerating the model inference.
The pruning schemes mainly consist: 
1) the flexible, unstructured pruning scheme that prunes weights at arbitrary locations \cite{han2016deep}, 
2) the regular, structured pruning that prunes whole filters/channels for CONV layers \cite{he2019filter}, 
and 3) the compiler-assisted, fine-grained structured pruning schemes~\cite{cai2020yolobile} that combining the high accuracy and high hardware parallelism.
Weight pruning provides a promising solution for accelerating inference of multi-person PE on mobile devices while preserving good accuracy.

Model architecture is another critical factor that significantly affects PE performance. For example, under similar computation FLOPs, the model depth has a direct impact on model accuracy and inference latency. Moreover, the unfriendly computation operators in a DNN model will introduce tremendous execution overhead on resource-limited model devices but usually has a minor impact on high-end server CPUs/GPUs. It is necessary to avoid using mobile-unfriendly operators for on-mobile PE models.

Besides the DNN model inference, the data post-processing is another essential part of the pose estimation process, which calculates and groups the keypoints of each individual obtained from the DNN inference results.
Different post-processing algorithms will significantly affect the end-to-end running time for the pose estimation.
Thus, choosing appropriate post-processing algorithms is also critical for both speed and accuracy consideration.

In this paper, to facilitate the fast and accurate multi-person PE on mobile devices, we propose a framework incorporating architecture optimization and weight pruning to find the desired
model architecture and pruning configurations to effectively accelerate inference of multi-person PE mobile devices.
\begin{itemize}
 \item We propose an optimization framework for the multi-person PE acceleration. The framework includes model architecture optimization and weight pruning to facilitate fast and accurate end-to-end PE on mobile devices.
 \item We propose several mobile-aware architecture optimization paradigms including model depth determination, mobile-unfriendly operator replacement, and post-processing selection.
 \item We explore the sensitivity of different function blocks and different layer types to the weight pruning and adopt suited pruning strategy for optimization.
 \item  We achieve significant acceleration for multi-person PE inference on smartphones while maintaining a high model accuracy.
 \item  Our achievements raise the promising future of various PE-related applications on mobile devices.
\end{itemize}

\section{Optimization Framework Design}
The objective of our framework is to provide an optimization flow to effectively accelerate PE models for fast inference on mobile devices while maintaining competitive model accuracy. Our framework mainly contains two parts, model architecture optimization, and weight pruning.

\subsection{Model Architecture Optimization}
\subsubsection{Model Depth Determination}
The models for multi-person PE tasks usually constitute several functional blocks. We use the OpenPose~\cite{openpose2019} model as an example.
The OpenPose model first uses a backbone model response for feature extraction.
The backbone depth will affect both final model accuracy and inference latency.
Generally, within a certain depth, a deeper backbone model provides a better final accuracy, but leads to a higher inference latency.
To avoid significantly increasing the model computation FLOPs, we increase the number of building blocks of the backbone model while shrinking the block width to maintain similar computation FLOPs.
As shown in Table~\ref{table1}, under similar computation FLOPs, a deeper version of the ResNet50 model with 72 layers provides a clear better final accuracy. 
But keep increasing model depth will further improve the accuracy.

Besides, to determine the number of cascaded refinement blocks is also important.
The accuracy improvement obtained by adding refinement blocks usually tends to be saturated after the first or second refinement block.
Thus, it is more economic to only add few refinement blocks, especially for an inference speed-driven PE design on mobile devices.

\subsubsection{Mobile-Unfriendly Operator Replacement}
Using computation-friendly operators is often not the top concern when designing networks pursuing high accuracy and using high-end server CPUs/GPUs. But it is critical for the implementation on mobile devices.
For example the Swish activation function used in EfficientNet requires exponential computation, which is mobile-unfriendly and significantly slows down the inference speed. This even makes EfficientNet slower than ResNet50, which requires higher computation FLOPs.
Thus, our framework replace the unfriendly operator with  more friendly alternative (e.g., hard-tanh).

Some special kinds of convolution (CONV) layers are inefficient in mobile inference such as the 7x7 CONV layer.
In Lightweight OpenPose~\cite{osokin2018lightweight_openpose}, they replace the 7x7 CONV layer with one 1x1 CONV layer followed by a 3x3 CONV layer and a dilation 3x3 CONV layer.
However, the dilation CONV layer requires access to non-consecutive input data, which leads to bad locality and hence a tremendous execution overhead.
In our framework, we replace the 7x7 CONV layer with three consecutive 3x3 CONV layers to preserve a similar receptive field as the 7x7 CONV layer but with a much higher speed.

\begin{table}[t]
    \renewcommand{\arraystretch}{1.0}
	\centering
    \begin{tabular}{ccccc}
    \hline
    Model & Num of layers & GFLOPs & AP & mAP\\
    \hline
    ResNet$50$ & 24 & 13.68 & 0.41 & 0.682\\
    ResNet$50_2$ & 36 & 16.05 & 0.423 & 0.69\\
    ResNet$50_3$ & 72 & 16.09 & 0.436 & 0.693\\
    ResNet$50_4$ & 147 & 16.46 & 0.433 & 0.69\\
    \hline
    \end{tabular}
    \caption{Accuracy under different depths of ResNet50 backbone.}
    \label{table1}
\end{table}

\subsubsection{Post-Processing Selection}
After the DNN model inference, all PE methods require a post-processing step.
There are several different post-processing methods, and different methods will result in significantly different post-processing times. Thus, selecting an appropriate post-processing method is an essential problem for PE implementations on mobile devices.
To have a fair comparison of post-processing time for different methods, we compared them using desktop CPU since some methods are mobile-unfriendly and hard to be applied on mobile devices.

In EfficientHRNet~\cite{efficienthrnet2020}, it proposes to use the way same as Associate Embedding, it needs to cluster the keypoints into multiple persons according to tags got from the output of the network and match the heatmaps for different resolutions, which is time-consuming.
Although the EfficientHRNet has smaller GFLOPs and higher accuracy, it takes around 2s for post-processing on desktop CPU and dominates the overall end-to-end inference time.
As for HRNet~\cite{hrnet2019}, it used the keypoint regression to improve keypoint detection and grouping, it still costs 0.2s.
In OpenPose~\cite{openpose2019}, it proposes to parse the heatmaps and PAFs obtained from the network output by performing a set of bipartite matchings
to associate body part candidates and assemble them into full-body poses for all people in the image. This post-processing takes about 0.02s.
Both the first two methods are not computational efficient and making the post-processing become the bottleneck for the latency. The PAF post-processing used in OpenPose is more desirable for the mobile-based implementation.

\subsection{Weight Pruning}

% \noindent\textbf{Pruning strategy:}
\paragraph{Pruning strategy:}
In order to maintain accuracy while achieving a higher acceleration, we use the fine-grained pruning schemes in our framework.
Since different sizes and types of DNN layers inherently have different redundancy, it is desirable to adopt a layer-wise pruning ratio.
Because sparse computation introduces execution overhead, for one layer, only with a prune ratio that exceeds a threshold can achieve actual acceleration. 
Our prune strategy for each layer is either prune with a relatively high ratio or not prune. 
We established a latency model to assist our prune ratio selection.

\paragraph{Sensitivity Analysis:}
As different functional blocks (e.g., backbone, heatmap branch, PAF branch) in a network exhibit different sensitivity to pruning, pruning on different functional blocks lead to different impacts on the final accuracy.
Therefore, we conducted a sensitivity analysis on different functional blocks in the model. We found that the backbone part has a higher tolerance for pruning and can be pruned more aggressively. Compared with the heatmap branch, the PAF branch is more sensitive to pruning, and the high pruning ratio will bring a relatively higher accuracy drop, so we adopt a milder pruning strategy for the PAF branch, that is, only prune a small number of layers or not prune.

\section{Results and Demonstrations}

\begin{table}[t]
\scriptsize
\sffamily
% \footnotesize\sffamily
    \centering
    \renewcommand{\arraystretch}{1.2}
\begin{adjustbox}{max width=1\columnwidth}
\begin{tabular}{@{} c c *{6}{c} @{c}}
\toprule
\bfseries 
\bfseries \begin{tabular}{@{}c@{}}Model\end{tabular}
& \bfseries \begin{tabular}{@{}c@{}}Input\\Size \end{tabular}
& \bfseries \begin{tabular}{@{}c@{}}Average\\Precision \end{tabular}
& \bfseries \begin{tabular}{@{}c@{}}Num of \\ Param. (M)\end{tabular}
& \bfseries \begin{tabular}{@{}c@{}}FLOPs\\ (G)\end{tabular}
& \bfseries \begin{tabular}{@{}c@{}}Friendly \\ Post-Proce.\end{tabular}
\\[2ex]
    % \toprule
    % \multicolumn{6}{c}{\textbf{Scale $\times 2$}} \\ 
    \midrule
    PersonLab & 1401 & 66.5 & 68.7 & 405.5 & $\times$ \\
    Associate Embedding & 512 & 65.5 & 138.9 & 222.3 & $\times$ \\
    HRNet & 512 & 64.4 & 28.5 & 38.9 & $\times$ \\
    HigherHRNet & 512 & 67.1 & 28.6 & 47.9 & $\times$ \\
    EfficientHRNet $H_{-4}$ & 384 & 35.7 & 3.7 & 2.1 & $\times$ \\
    EfficientHRNet $H_{-3}$ & 416 & 44.8 & 6.9 & 4.2 & $\times$ \\
    % \midrule
    % \multicolumn{6}{c}{\textbf{Scale $\times 4$}} \\ 
    \midrule
    OpenPose & 368 & 48.6 & 52.3 & 136.1 & $\surd$ \\
    Lightweight OpenPose & 368 & 40 & 4.1 & 9 & $\surd$ \\
    \midrule
    Ours ResNet$50_3$ (dense) & 368 & 43.9 & 6.77 & 16.4 & $\surd$ \\
    Ours ResNet$50_3$ (1.80$\times$) & 368 & 42.0 & 3.76 & 9.7 & $\surd$ \\
    Ours ResNet$50_3$ (2.15$\times$) & 368 & 41.5 & 3.15 & 8.3 & $\surd$ \\
    Ours EfficientNet (2.01$\times$) & 368 & 41.6 & 1.68 & 4.7 & $\surd$ \\
    \bottomrule

\end{tabular}
\end{adjustbox}
    \caption{Comparison with state-of-the-art pose estimation methods.}
        \label{table:result}
\end{table}

\begin{figure}[t]
     \centering
     \includegraphics[width=0.8\columnwidth]{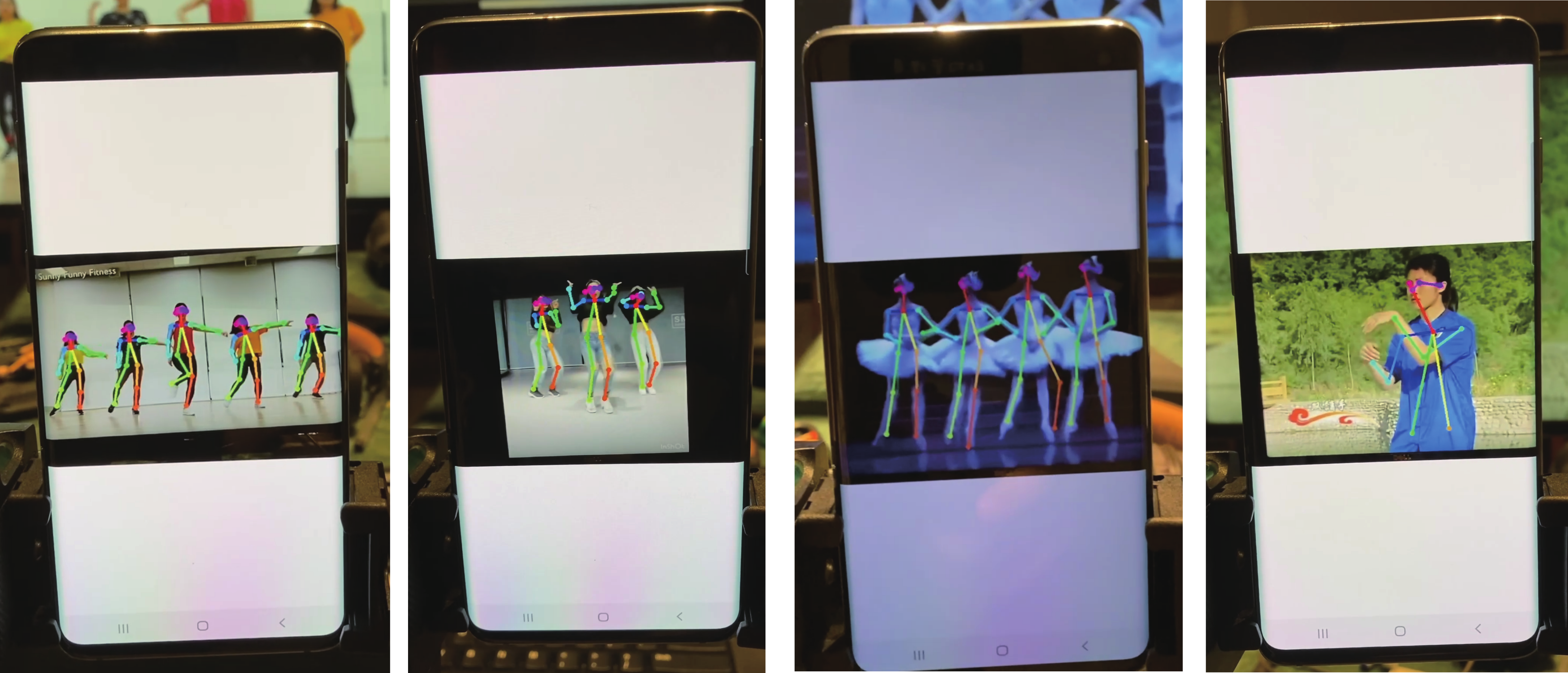}  
     \caption{Demonstration of multi-person pose estimation on Samsung Galaxy S10 smartphone.}
    \label{fig:demo} 
\end{figure}

\begin{figure}[t]
     \centering
     \includegraphics[width=0.8\columnwidth]{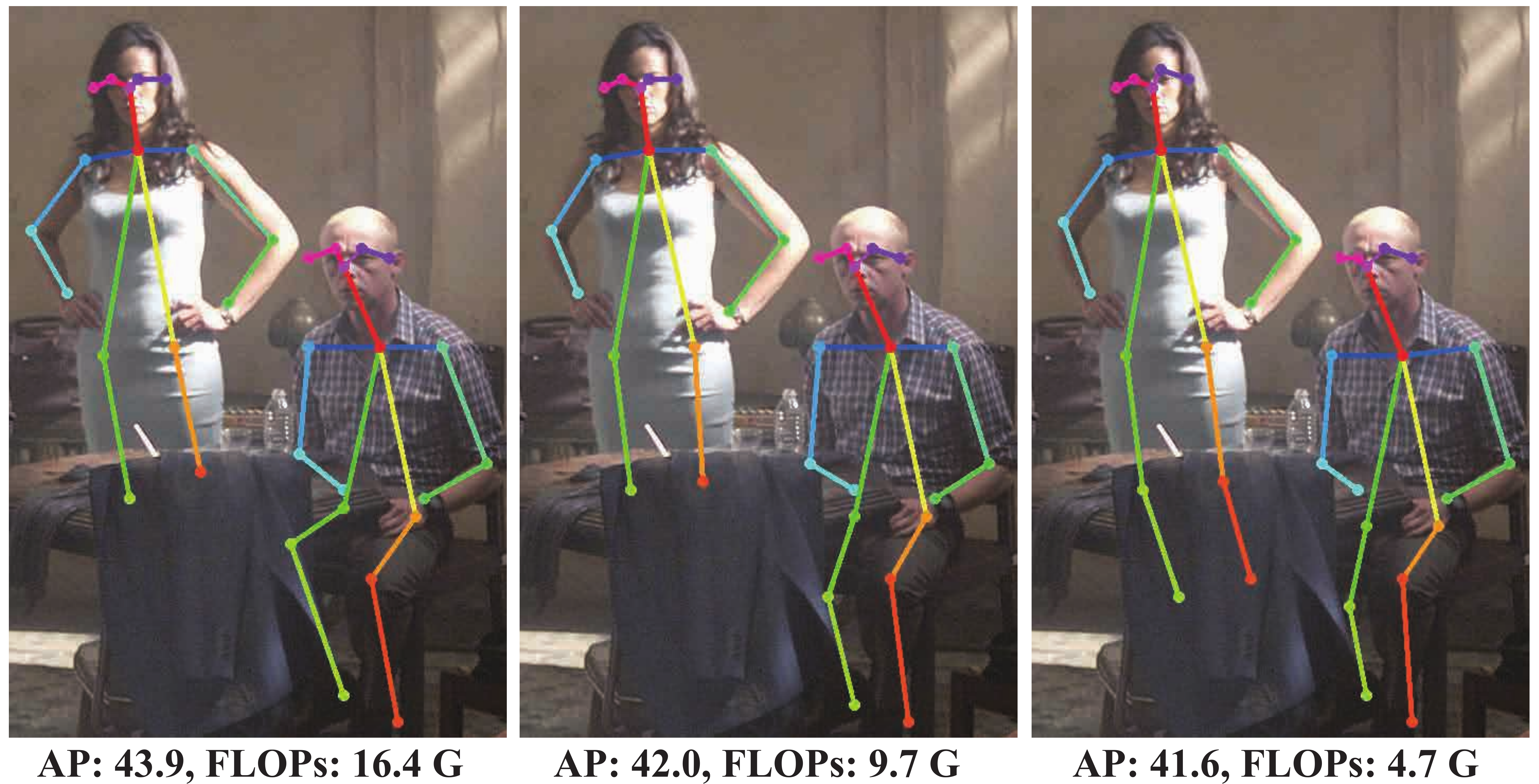}  
     \caption{Visual quality of multi-person pose estimation using our optimized models.}
    \label{fig:visual}  
\end{figure}

\begin{figure}[t]
     \centering
     \includegraphics[width=0.8\columnwidth]{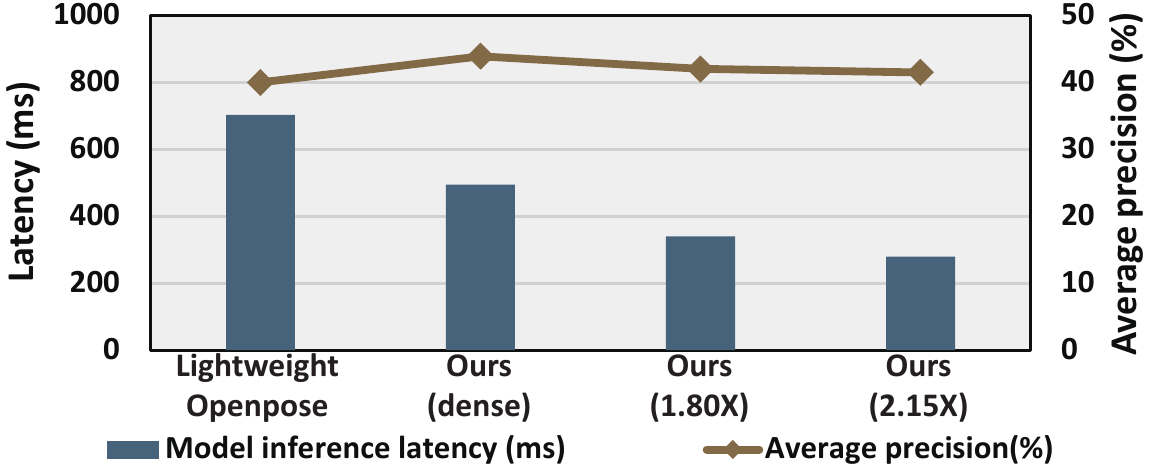}
     \caption{Model inference latency on Samsung Galaxy S10.}
    \label{fig:latency}  
\end{figure}

\paragraph{Experiment Setup:}
We trained our models on MS COCO dataset~\cite{COCOdata} using PyTorch API. 
The model inference latency is measured on a Samsung Galaxy S10 smartphone using a mobile GPU.
Figure~\ref{fig:demo} shows the demonstrations of multi-person pose estimation on the smartphone.
We follow the design methodology of OpenPose~\cite{openpose2019} since the PAF the most mobile-friendly post-processing compared to others.
Our optimizations include model depth determination, mobile-unfriendly operator replacement, and weight pruning.
Note that our optimization framework is general and can also be used to optimize other design methodologies.

\paragraph{Accuracy \& Visual Quality:}
We compared our optimized models with representative PE methods~\cite{2018personlab,associateembedding2017,hrnet2019,higherhrnet2020,efficienthrnet2020,openpose2019,osokin2018lightweight_openpose},
% in average precision (AP), number of parameters, and Computation FLOPs
as shown in Table~\ref{table:result}.
Most of the methods either have huge computation FLOPs or mobile-unfriendly post-processing, or both, making them not applicable to on-mobile PE tasks.
When compare with the Lightweight OpenPose (LWOP), which has the potentiality to be deployed on mobile devices, our optimized models achieve higher accuracy and lower FLOPs.
Figure~\ref{fig:visual} shows the visual quality of multi-person PE using our different optimized models.
As shown in the figure, we can maintain high visual quality with a significant reduction in computation FLOPs.

\paragraph{Inference Latency:}
Figure~\ref{fig:latency} shows the model inference latency comparison of our optimized models and LWOP model on smartphones. As mentioned in Table~\ref{table:result} the LWOP model has smaller computation FLOPs compared to our dense model and 1.80$\times$ pruned model (i.e., 9G vs. 16.4G and 9.7G), but our optimized models are 1.42$\times$ and 2.07$\times$ faster, respectively. This is because we replaced the mobile-unfriendly layers in the model. And our best model achieves 2.51$\times$ speedup with higher accuracy compared to LWOP.

\section{Conclusion}
We propose an optimization framework towards enabling fast and accurate multi-person pose estimation on mobile devices. Our optimized models achieve both higher accuracy and faster inference speed on mobile device.
% compared to representative lightweight pose estimation design.

\section*{Acknowledgements}
This project is partly supported by National Science Foundation (NSF) under CNS-1739748, Army Research Office/Army Research Laboratory via grant W911NF-20-1-0167 (YIP) to Northeastern University, a grant from Semiconductor Research Corporation (SRC), and Jeffress Trust Awards in Interdisciplinary Research.
% Any opinions, findings, and conclusions or recommendations expressed in this material are those of the authors and do not necessarily reflect the views of NSF, ARO/ARL, SRC, or Thomas F. and Kate Miller Jeffress Memorial Trust. 

% funded by the National Science Foundation Awards CCF-1937500 and CNS-1909172.

%% The file named.bst is a bibliography style file for BibTeX 0.99c
{
\bibliographystyle{named}
\bibliography{ijcai21}

\begin{thebibliography}{}

\bibitem[\protect\citeauthoryear{Alejandro~Newell}{2017}]{associateembedding2017}
Jia~Deng Alejandro~Newell, Zhiao~Huang.
\newblock Associative embedding: Endto-end learning for joint detection and
  grouping.
\newblock In {\em Advances in Neural Information Processing Systems (NIPS)},
  2017.

\bibitem[\protect\citeauthoryear{Bowen~Cheng}{2020}]{higherhrnet2020}
Jingdong Wang Honghui Shi Thomas S. Huang Lei~Zhang Bowen~Cheng, Bin~Xiao.
\newblock Higherhrnet: Scale-aware representation learning for bottom-up human
  pose estimation.
\newblock In {\em Proceedings of the IEEE Conference on Computer Vision and
  Pattern Recognition (CVPR)}, 2020.

\bibitem[\protect\citeauthoryear{Cai \bgroup \em et al.\egroup
  }{2020}]{cai2020yolobile}
Yuxuan Cai, Hongjia Li, Geng Yuan, Wei Niu, Yanyu Li, Xulong Tang, Bin Ren, and
  Yanzhi Wang.
\newblock Yolobile: Real-time object detection on mobile devices via
  compression-compilation co-design.
\newblock {\em arXiv preprint arXiv:2009.05697}, 2020.

\bibitem[\protect\citeauthoryear{{Cao} \bgroup \em et al.\egroup
  }{2019}]{openpose2019}
Z.~{Cao}, G.~{Hidalgo Martinez}, T.~{Simon}, S.~{Wei}, and Y.~A. {Sheikh}.
\newblock Openpose: Realtime multi-person 2d pose estimation using part
  affinity fields.
\newblock {\em IEEE Transactions on Pattern Analysis and Machine Intelligence},
  2019.

\bibitem[\protect\citeauthoryear{Christopher~Neff}{2020}]{efficienthrnet2020}
Steven Furgurson Hamed~Tabkhi Christopher~Neff, Aneri~Sheth.
\newblock Efficienthrnet: Efficient scaling for lightweight high-resolution
  multi-person pose estimation.
\newblock {\em arXiv preprint arXiv:2007.08090}, 2020.

\bibitem[\protect\citeauthoryear{Fang \bgroup \em et al.\egroup
  }{2017}]{fang2017rmpe}
Hao-Shu Fang, Shuqin Xie, Yu-Wing Tai, and Cewu Lu.
\newblock {RMPE}: Regional multi-person pose estimation.
\newblock In {\em ICCV}, 2017.

\bibitem[\protect\citeauthoryear{Han \bgroup \em et al.\egroup
  }{2016}]{han2016deep}
Song Han, Huizi Mao, and William~J. Dally.
\newblock Deep compression: Compressing deep neural networks with pruning,
  trained quantization and huffman coding.
\newblock In {\em International Conference on Learning Representations (ICLR)},
  2016.

\bibitem[\protect\citeauthoryear{He \bgroup \em et al.\egroup
  }{2019}]{he2019filter}
Yang He, Ping Liu, Ziwei Wang, Zhilan Hu, and Yi~Yang.
\newblock Filter pruning via geometric median for deep convolutional neural
  networks acceleration.
\newblock In {\em Proceedings of the IEEE Conference on Computer Vision and
  Pattern Recognition (CVPR)}, pages 4340--4349, 2019.

\bibitem[\protect\citeauthoryear{Jingdong~Wang and Xiao}{2019}]{hrnet2019}
Tianheng Cheng Borui Jiang Chaorui Deng Yang Zhao Dong Liu Yadong Mu Mingkui
  Tan Xinggang Wang Wenyu~Liu Jingdong~Wang, Ke~Sun and Bin Xiao.
\newblock Deep highresolution representation learning for human pose
  estimation.
\newblock {\em IEEE transactions on pattern analysis and machine intelligence
  (T PAMI)}, 2019.

\bibitem[\protect\citeauthoryear{Li \bgroup \em et al.\egroup
  }{2021a}]{li2021real}
Hongjia Li, Geng Yuan, Wei Niu, Yuxuan Cai, Mengshu Sun, Zhengang Li, Bin Ren,
  Xue Lin, and Yanzhi Wang.
\newblock Real-time mobile acceleration of dnns: From computer vision to
  medical applications.
\newblock In {\em 2021 26th Asia and South Pacific Design Automation Conference
  (ASP-DAC)}, pages 581--586. IEEE, 2021.

\bibitem[\protect\citeauthoryear{Li \bgroup \em et al.\egroup
  }{2021b}]{li2021npas}
Zhengang Li, Geng Yuan, Wei Niu, Pu~Zhao, Yanyu Li, Yuxuan Cai, Xuan Shen,
  Zheng Zhan, Zhenglun Kong, Qing Jin, Zhiyu Chen, Sijia Liu, Kaiyuan Yang, Bin
  Ren, Yanzhi Wang, and Xue Lin.
\newblock Npas: A compiler-aware framework of unified network pruning and
  architecture search for beyond real-time mobile acceleration.
\newblock In {\em Proceedings of the IEEE Conference on Computer Vision and
  Pattern Recognition (CVPR)}, 2021.

\bibitem[\protect\citeauthoryear{Niu \bgroup \em et al.\egroup
  }{2020}]{niu2020achieving}
Wei Niu, Zhenglun Kong, Geng Yuan, Weiwen Jiang, Jiexiong Guan, Caiwen Ding,
  Pu~Zhao, Sijia Liu, Bin Ren, and Yanzhi Wang.
\newblock Achieving real-time execution of transformer-based large-scale models
  on mobile with compiler-aware neural architecture optimization.
\newblock {\em arXiv preprint arXiv:2009.06823}, 2020.

\bibitem[\protect\citeauthoryear{Osokin}{2018}]{osokin2018lightweight_openpose}
Daniil Osokin.
\newblock Real-time 2d multi-person pose estimation on cpu: Lightweight
  openpose.
\newblock In {\em arXiv preprint arXiv:1811.12004}, 2018.

\bibitem[\protect\citeauthoryear{Papandreou \bgroup \em et al.\egroup
  }{2018}]{2018personlab}
George Papandreou, Tyler Zhu, Liang-Chieh Chen, Spyros Gidaris, Jonathan
  Tompson, and Kevin Murphy.
\newblock Personlab: Person pose estimation and instance segmentation with a
  bottom-up, part-based, geometric embedding model.
\newblock In {\em Proceedings of the European Conference on Computer Vision
  (ECCV)}, pages 269--286, 2018.

\bibitem[\protect\citeauthoryear{Tsung-Yi~Lin}{2014}]{COCOdata}
Serge Belongie Lubomir Bourdev Ross Girshick James Hays Pietro Perona Deva
  Ramanan C. Lawrence Zitnick Piotr~Dollár Tsung-Yi~Lin, Michael~Maire.
\newblock Microsoft coco: Common objects in context.
\newblock In {\em Proceedings of the European Conference on Computer Vision
  (ECCV)}, 2014.

\bibitem[\protect\citeauthoryear{Zhang \bgroup \em et al.\egroup
  }{2021}]{zhang2021clicktrain}
Chengming Zhang, Geng Yuan, Wei Niu, Jiannan Tian, Sian Jin, Donglin Zhuang,
  Zhe Jiang, Yanzhi Wang, Bin Ren, Shuaiwen~Leon Song, et~al.
\newblock Clicktrain: Efficient and accurate end-to-end deep learning training
  via fine-grained architecture-preserving pruning.
\newblock In {\em The 35th ACM International Conference on Supercomputing (ICS
  2021)}, 2021.

\bibitem[\protect\citeauthoryear{Zhao \bgroup \em et al.\egroup
  }{2020}]{zhao2020achieving}
Pu~Zhao, Wei Niu, Geng Yuan, Yuxuan Cai, Hsin-Hsuan Sung, Wujie Wen, Sijia Liu,
  Xipeng Shen, Bin Ren, Yanzhi Wang, et~al.
\newblock Achieving real-time lidar 3d object detection on a mobile device.
\newblock {\em arXiv preprint arXiv:2012.13801}, 2020.

\end{thebibliography}
}

\end{document}